\documentclass{article}

\usepackage{arxiv}
\usepackage{CJKutf8}
\usepackage[pdftex]{graphicx}
\usepackage {multirow}
\usepackage {booktabs}
\usepackage {amssymb}
\usepackage {amsmath}
\usepackage[section]{placeins}

\usepackage[utf8]{inputenc} 
\usepackage[T1]{fontenc}    
\usepackage{hyperref}       
\usepackage{url}            
\usepackage{booktabs}       
\usepackage{amsfonts}       
\usepackage{nicefrac}       
\usepackage{microtype}      
\usepackage{cleveref}       
\usepackage{lipsum}         
\usepackage{graphicx}
\usepackage{natbib}
\usepackage{doi}

\title{Evaluating and Enhancing Large Language Models for Conversational Reasoning on Knowledge Graphs}

\date{}
\author{ Yuxuan Huang \\
	College of Software \\ 
    Jilin University \\
    Changchun, China \\
	\texttt{airoura@163.com}
 }

\renewcommand{\headeright}{}
\renewcommand{\undertitle}{}
\renewcommand{\shorttitle}{\textit{}}

\hypersetup{
pdftitle={Evaluating and Enhancing Large Language Models for Conversational Reasoning on Knowledge Graphs},
pdfsubject={cs.CL, cs.AI},
pdfauthor={Yuxuan Huang},
pdfkeywords={Knowledge Graph, Large Language Model, Deep Reinforcement Learning},
}

\begin{document}
\maketitle
\begin{abstract}
The development of large language models (LLMs) has been catalyzed by advancements in pre-training techniques. These models have demonstrated robust reasoning capabilities through manually designed prompts. In this work, we evaluate the conversational reasoning capabilities of the current state-of-the-art LLM (GPT-4) on knowledge graphs (KGs). However, the performance of LLMs is constrained due to a lack of KG environment awareness and the difficulties in developing effective optimization mechanisms for intermediary reasoning stages. We further introduce LLM-ARK, a LLM grounded KG reasoning agent designed to deliver precise and adaptable predictions on KG paths. LLM-ARK leverages Full Textual Environment (FTE) prompt to assimilate state information within each reasoning step. We reframe the challenge of multi-hop reasoning on the KG as a sequential decision-making task. Utilizing the Proximal Policy Optimization (PPO) online policy gradient reinforcement learning algorithm, our model is optimized to learn from rich reward signals. Additionally, we conduct an evaluation of our model and GPT-4 on the OpenDialKG dataset. The experimental results reveal that LLaMA-2-7B-ARK outperforms the current state-of-the-art model by 5.28 percentage points, with a performance rate of 36.39\% on the target@1 evaluation metric. Meanwhile, GPT-4 scored 14.91\%, further demonstrating the effectiveness of our method.
\end{abstract}

\section{Introduction}
\label{sec:introduction}
With significant progress in large language models (LLMs), researchers have recognized their superior capabilities in the field of natural language processing (NLP)\citep{liu2023comprehensive,shakarian2023independent,lai2023chatgpt}. Reasoning ability stands as the most demonstrative of AI intelligence. Recently, to boost the performance of LLMs in reasoning tasks, we noted various optimization strategies adopted by researchers such as Chain of Thought (COT)\citep{wei2023chainofthought} and decomposing subtasks\citep{kazemi2023lambada}. Currently, the reasoning method of LLMs have received limited attention in the conversational KG reasoning task. This research aims to address this gap in the field.

\begin{figure}[ht]
    \label{figure1}
    \begin{center}
        \includegraphics[width=1\linewidth]{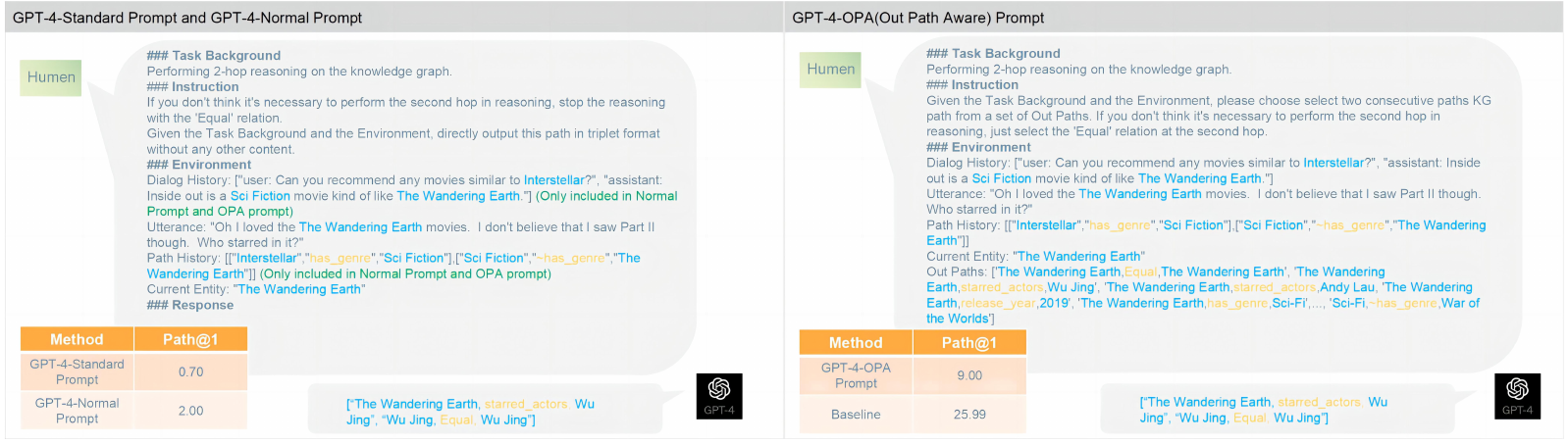}
    \end{center}
    \caption{Manual prompts on the OpenDialKG dataset. Compare to GPT-4-Standard, GPT-4-Normal has more awareness of dialog context and path history, while GPT-4-OPA has more awareness of 2-hop exit path subgraphs compared to the former two. The experimental results show that the more environmental information GPT-4 perceives, the higher the knowledge graph reasoning path@1 evaluation metric score.}
\end{figure}

Knowledge Graphs composed of vertices or entities connected by edges or relations, gaining popularity in knowledge-based dialogue systems for its structured disposition. Conversational reasoning models are able to traverse the KG based on conversational context to introduce diverse entities and attributes to make replies more engaging, as well as to improve the logic of the model to mitigate illusions\citep{rawte-etal-2023-troubling, dong2022faithful}. Previous work\citep{moon-etal-2019-opendialkg, zhang-etal-2020-grounded, Ni_Pandelea_Young_Zhou_Cambria_2022, tuan-etal-2022-towards} mainly relied on supervised learning methods. To assess the capabilities of current SOTA LLM: GPT-4\citep{openai2023gpt4}, we initially examine the proficiency of LLMs on KG reasoning, as illustrated in Figure \ref{figure1}, determining their potential application in the KG domain. Empirical studies reveal that, despite demonstrating reasonable performance on KG tasks, indicative of their proficiency in managing complex problems, understanding contextual relationships, and utilizing pre-training knowledge, LLMs still present issues and fall short when compared with state-of-the-art models.

There are two main challenges applying LLM-based agents. On the one hand, LLMs suffer from a limited perception of variable reasoning environments. The alignment between LLMs’ knowledge and the environment can be wrong and limit functional competence due to lack of grounding\citep{carta2023grounding}. If properly grounded, the model's structure would be both simplified and effective. For KG reasoning tasks, as shown in Figure \ref{figure1}, the agent achieved better scores when provided with as much information about the environment as possible, such as dialog history, inference path history, and all exit paths. Although LLMs are not designed to take actions, \citet{peng2023check, carta2023grounding, pmlr-v202-du23f} found that it can be achieved good results in downstream decision making by simply feeding full textual representations as inputs to LLMs.

On the other hand, \citet{yao2023retroformer} indicate that there is a lack of systematic methods for consistent model refinement. In essence, LLMs fall short in possessing essential mechanisms for optimizing intermediary reasoning processes in multi-hop reasoning tasks. This is mainly attributed to the fact that manual prompt tuning is widely used in many application scenarios. It has been observed that LLM-based agents can easily fall into infinite loops if state is not handled properly, and inevitably run into prompt length problems when the trajectory becomes longer. In addition, the design of prompt is also a challenge because an entity may have more than 100 exit edges, all of which are formatted into prompt which is impractical in a knowledge graph environment. LLMs often encounter these issues because they are not designed or trained for action-agent applications.

We introduce \textbf{LLM-ARK}, an effective framework that employs \textbf{LLM} as an \textbf{Agent} for \textbf{R}easoning on \textbf{K}nowledge Graphs. We employ LLMs as agent and express the Large model KG inference task as a reinforcement learning sequential decision-making problem, and using a Full-Textual-Environment prompt to aggregate multiscale inputs. Moreover, our agent architecture does not necessitate access to LLM parameters or gradient propagation through it. Instead, we adopt a policy gradient approach where the Actor LLM functions as part of the environment. This configuration enables the model to learn from diverse reward signals across varied tasks and environments. In summary, our contributions are as follows: 
\begin{itemize}
    \item  We assess the capabilities of state-of-the-art LLM: GPT-4, on large-scale KG inference datasets and analyze the experimental results in detail to understand the causes of their inferior performance. 
    \item  To enhance the performance of the LLM agents, we introduce LLM-ARK. Our method expresses the KG dialog inference problem as a reinforcement learning sequential decision-making issue, using a Full-Textual-Environment prompt to aggregate multiscale inputs, dual-environment sensing on the state and decision side and leverage LLMs to explore on KGs.
    \item  Furthermore, we update only the parameters of the PA-MLP that are part of the our agent using the policy gradient method, freezing the parameters of the LLM. This approach enables learning from diverse reward signals during interactions with the environment and improves the efficiency of training.
\end{itemize}

\section{Related Work}
\label{sec:related work}
\subsection{KG Reasoning on Dialog Systems}
Given its structured nature, Knowledge Graphs are becoming an increasingly popular external information source in knowledge-based systems. \citet{moon-etal-2019-opendialkg} developed a retrieval system designed to generate responses based on a graph reasoning task. They employed a graph walker to navigate the graph, propelled by the symbol transformation conditions of the dialog context. \citet{jung-etal-2020-attnio} utilizing graph attention techniques to navigate the conditional graph of a conversation within a KG dialogue system. The model computes an incoming attention fow to represent entities and an outgoing attention fow to select KG paths. However, this approach cannot be extended to long KG path prediction due to the exponential increase in computational complexity. \citet{Ni_Pandelea_Young_Zhou_Cambria_2022} introduced a hierarchical reinforcement learning KG inference model that aggregates multiple inputs utilizing an attention mechanism. This approach instructs the model to reason in one step and then fine-tunes it using a goal-directed reinforcement learning. \citet{tuan-etal-2022-towards} employed a single transformer model that walks directly over large-scale KGs, reasoning over fine-tunable KGs to generate responses. Similarly, \citet{luo2023reasoning} initially create relational paths derived from KGs as high-confidence plans, which are later utilized to extract valid reasoning paths from KGs for confident reasoning. \citet{sun2023thinkongraph} leverage KGs to augment LLMs for deep and responsible reasoning. The framework explores and infers by identifying entities relevant to a given question and retrieving relevant triples from external KGs. This iterative process generates multiple inference paths until enough information is gathered to answer the question or maximum depth is reached.

\subsection{LLMs with Reinforcement Learning}
Reinforcement learning and large models are divided into two main aspects of the combination, the first aspect further improves the ability of LLM to understand and follow user instructions through reinforcement learning based methods\citep{Ouyang2022TrainingLM}. \citet{yao2023retroformer} employed a special RLHF technique to tailor the model to human preferences, generating beneficial, non-toxic, and safe data for training while also training reward models to evaluate LLMs. Retroformer, a significant improvement over Chain of Thought (COT), is primarily applied to reasoning tasks and uses a unique RLHF method. \citet{shinn2023reflexion} introduce a novel framework, Reflexion, that strengthens linguistic agents through linguistic feedback, rather than updating weights. The Reflexion agent verbally reflects on task feedback signals, and then stores its reflection text in an episodic memory buffer to make better decisions in subsequent trials.

The second aspect is to further improve the applicability of LLM on real-world tasks through a reinforcement learning-based approach, since training a LLM public NLP task/dataset can only cover a small portion of the real world, and reinforcement learning can train LLM-based intelligences to explore the realization of various real-world goals. \citet{carta2023grounding} studied LLMs interaction with physical environments. Using an interactive textual environment designed to study a series of spatial and navigational tasks and using online reinforcement learning to improve its performance to solve goals. \citet{huang2022inner} consistently integrate feedback from diverse sources into the planning language cues of the LLM, thereby enabling it to reason and replan to solve complex problems in both simulated and real-world environments. \citet{singh2022progprompt} propose a procedural LLM hint structure that facilitates plan generation functionality in contextual environments, robot capabilities, and tasks.

\section{Methods}
\label{sec:methods}
\subsection{Overview}
\begin{figure*}[!htbp]
    \begin{center}
        \includegraphics[width=1\linewidth]{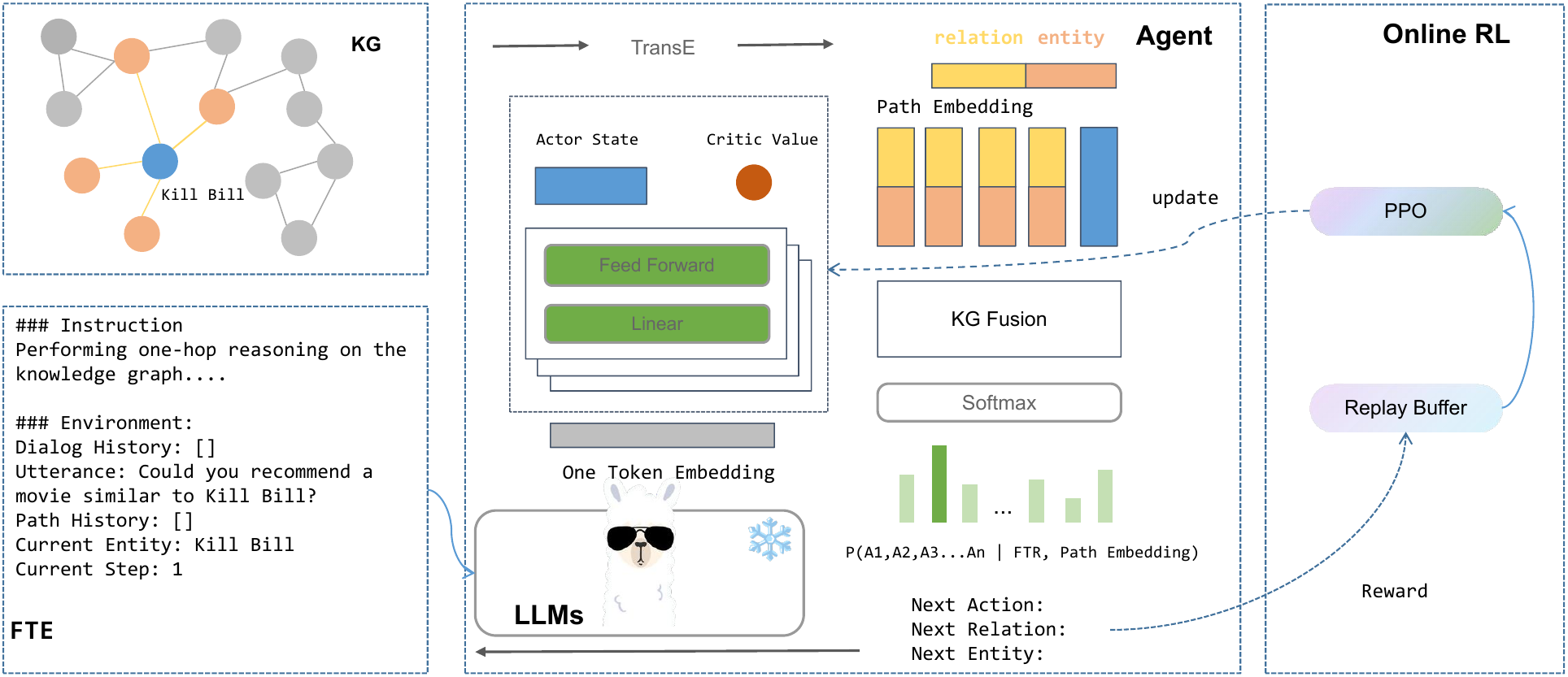}
    \end{center}
    \caption{The overall architecture of LLM-ARK.}
\label{figure2}
\end{figure*}
As shown in Figure \ref{figure2}, our model has the following main components, FTE (Full-Textual-Environment), LLM (Large Language Model) and RL(Reinforcement Learning). FTE can be seen as state manager, using a Full-Textual-Environment prompt to aggregate multi-scale inputs, updating and maintaining state transfers between itself and the environment. At first LLMs obtain a richly informative representation of state embeddings. To capture the path embedding information of the KG, we pre-train the KG on TransE\citep{fan-etal-2014-transition}. Rather than directly introducing the probability distribution of the action space, our Actor feeds the probability distribution along with the path embedding, subsequently eliminating invalid paths (we utilise 'Pad' for this adaptation process) before outputting a precise and legitimate action. We formulate the large model KG inference task within an online reinforcement learning framework and continuously optimize the decision network based on the collected experience in replay buffer. Finally, we refine the adapter using the Proximal Policy Optimization (PPO) online reinforcement learning method. In this section, we will first describe the method used to evaluate GPT-4, and then present each of the modules of our model in turn.

\subsection{Manual Prompt Tuning}
As illustrated in Figure \ref{figure1}, we detail the prompting schemes, encompassing the standard prompt, normal prompt and out path aware prompt. To guide LLMs in performing specific dialogue tasks, we can formulate the standard prompt and normal prompt scheme as:
\begin{equation}
    \begin{aligned}
        p(r|D, C)
    \end{aligned}
\end{equation}
Given the task background D and the conversation history C, instruct the LLM to generate the response r. More complex path aware prompt aims to provide alternative options for LLMs to decide what kinds of actions should be taken in the response, instead of simply responding to the instruction. It can be formulated as:
\begin{equation}
    \begin{aligned}
        p(a, r|D, C, A)
    \end{aligned}
\end{equation}
Given the task background D, conversation history C, and a set of potential dialogue acts A, the LLM is guided to select the most appropriate dialogue act $a \in $A, which then generates the response r.

\subsection{LLM-ARK}
Knowledge Graphs are structured knowledge networks composed of vertices, interpreted as entities, associated via edges or relationships. Let $\mathcal{E}$ stand for a collection of entities and $\mathcal{R}$ for a collection of relations. We represent the external KG as $\mathcal{G}$ = \{V,E,$\mathcal{R}$\}, where V and E denote the vertices and edges of the graph, respectively. Note that V = $\mathcal{E}$ and E $\subseteq V $ × $ \mathcal{R}$ × V. Let $v$ denote a node and $e$ denote an edge in $\mathcal{G}$. Given dialog context $X = $ and $\mathcal{G}$, we can identify an entity in the KG (e.g., an entity name The Wondering Earth) and and represent it as $v_s, v_s \in V$. The goal is to select a proper edge $e_t$ at the t-th timestamp for one-hop reasoning.

Graph attention-based models require significant annotation effort since all potential paths must be evaluated, which can be computationally expensive for large Knowledge Bases (KBs) with millions of entities. To overcome this challenge, our study employs a policy gradient model that efficiently traverses the KG to select relationships and ultimately achieves the target, demonstrating proficiency in multi-hop reasoning. 

KG reasoning naturally reduces to a finite horizon, deterministic partially observed Markov decision process that lie on a KG $\mathcal{G}$. We formulate KG reasoning as a Markov Decision Process (MDP) described by a six-tuple $(S, O, A, T, R, \gamma)$:

\begin{itemize}
    \item State. $S$ is an infinite set of environment states, which encode information stored in Working Memory, including task background $tb$, user query $q$, dialog history $h$, current entity $v_c$, path history $ph$, current step $t$, The normal state is represented using a six-tuple: $S = (tb, q, h, v_c, ph, t)$.
    \item Observation. The complete state of the environment can be observed. Formally, the observation function $O = S$.
    \item Action. The set of possible actions $\mathcal{A}$ from a state $S$ consists of all the environment information. Formally $ \mathcal{A}_s = \{e \in E \colon S \} \cup \{(s,\varnothing,s)\}$. This means that each state's agent can choose one of all output edges of the current entity.
    \item Transition. Depending on the edge selected by the agent at time step $t$, the environment is changed deterministically by updating the state to the new environment. For single turn dialogue, we update current entity, path history and step.
    Formally, the transition function $\colon \mathcal{\delta \colon S \times A \rightarrow S}$ is defined by $\mathcal{\delta}(S, A) = (tb, q, h, v_c^{'}, ph^{'}, t^{'}) $,
    For multi-turn dialogue wo also need to update user query and dialogue history.
    $\mathcal{\delta}(S, A) = (tb, q^{'}, h^{'}, v_c^{'}, ph^{'}, t^{'}) $,
    where $S = (tb, q, h, v_c, ph, t)$.
    \item Reward. We have a final reward of +1 if the current entity is the target entity $v_g$ and -1 otherwise. if $S_t = (tb, q^{'}, h^{'}, v_c^{'}, ph^{'}, k^{'})$ is the final state, then we have a final reward of +1 if $v_c^{'} = v_g$ , else -1.
    \item $\gamma$ denote reward discounts factor are used to compute the reward information of each intermediate process when agent reaches the goal, or the end of the maximum step $t$.
\end{itemize}

\subsubsection{Full Textual Environment}
This module tracks the agent's state that captures all essential information in the conversation so far. FTE is a text dictionary structure, the same as Prompt Engineering's normal prompt format.

\subsubsection{Agent}
Inference to previous work \citet{carta2023grounding}, we use standard RL practices by adding action heads - a Multi-Layer Perceptron (MLP) on top of the LLM. Thus, we can use only pretrained operations from the LLM and leverage language modeling heads’ prior, this method is robust to any action space and can thus be used on any textual environment with no change. Agent has two components: LLM and PA-MLP.

\textbf{LLM}
We initially utilize a LLM to encode the state $S$ into a continuous vector $s \in \mathcal{\mathbb{R}}^{2d}$. 
As a rule of thumb, for BERT models the cls token is used to represent the semantics of the whole sequence, while standard transformers and GPT-like LLMs use the embedding of the last token. We used the model on huggingface hub as well as the code to get the sequence vector representation\footnote{\url{https://huggingface.co}}.
It is defined by:
\begin{equation}
    \begin{aligned}
        s = llm(fte)
    \end{aligned}
\end{equation}
\textbf{PA-MLP}
Instead of just adding an MLP with a single output for the value on top of the last layer of the first decoder block as in the conventional multicategorization task, to enhance the ability of the large model to perceive the environment, we further fused the hidden state after the MLP with the Knowledge Graph exit path information, called PA-MLP (\textbf{P}ath \textbf{A}ware MLP).
Recall that each possible action represents an outgoing edge $e$ with information about the edge relation label $r_l$ and the target vertex/entity $v_d$. So the embedding for each $A_{t}$ is $[r_l ; v_d]$, and stacking the embeddings for all outgoing edges we get the matrix $A_t$. The network taking this as input is parameterized as a three-layer feed-forward network (MLP) with $\tanh$ nonlinearity, which takes the FTE representation $s$ and the embedding for the outgoing paths embedding and outputs a probability distribution over the possible actions from which a discrete action is sampled. The dimension of the MLP output hidden state is equal to the dimension of the path embedding. Finally, formulated as:
\begin{equation}
    \begin{aligned} \mathbf{h}_{\mathbf{t}} &=\mathbf{A}_{\mathbf{t}}\left(\mathbf{W}_{\mathbf{3}}\left(\operatorname{tanh}\left(\mathbf{W}_{\mathbf{2}} \left(\operatorname{tanh}\left(\mathbf{W}_{\mathbf{1}}\left(s_{t}\right)\right)\right)\right)\right)\right) \\ a_{t} & \sim \text { Categorical }\left(\operatorname{softmax}\left(\mathbf{h}_{\mathbf{t}}\right)\right) \end{aligned}
\label{EQUATION-PA-MLP-1}
\end{equation}

\subsubsection{Training}

\textbf{Optimizer}
Our model is optimized by utilizing the experience accumulated by agent during KG reasoning. More formally, for the above policy network ($\pi_\theta$), we want to find the parameter $\theta$ that maximizes the reward.

\begin{equation}
    \begin{split}
        J(\theta)=\mathbb{E}_{\left(e_{s}, \mathcal{P}, e_{g}\right) \sim D} \mathbb{E}_{A_{1}, \ldots, A_{T-1} \sim \pi_{\theta}} \\
        \left [R\left(S_{t}\right) \mid S_{1}=\left(s_{1}\right)\right], 
    \end{split}
\end{equation}
where we assume that there is a true underlying distribution $(e1,r, e2) \sim \mathcal{P} $. To address this optimization challenge, we adopt an online reinforcement learning policy gradient algorithm, Proximal Policy Optimization (PPO). PPO is a family of policy optimization methods that use multiple epochs of stochastic gradient ascent to perform each policy update. These methods have the stability and reliability of trust-region methods\citep{DBLP:journals/corr/SchulmanWDRK17}.
For value approximation, we include a three-layer feed-forward network with a single output for the value, given by:
\begin{equation}
    \begin{aligned} \mathbf{V}
    &=\mathbf{W}_{\mathbf{3}} \left(\operatorname{tanh}\left(\mathbf{W}_{\mathbf{2}} \left(\operatorname{tanh}\left(\mathbf{W}_{\mathbf{1}}\left(s_{t}\right)\right)\right)\right)\right)
    \end{aligned}
\end{equation}
Significantly, the LLM remains frozen for both the actor and critic modules, with only the linear forward layer being trained.

\textbf{Replay Buffer}
The replay buffer stores the triplets $rb = (v_c, s, logit, a, s^{'}, done)$ of the reflection prompt, indicating the current entity, the current state, logits, the selected action, the next state, and whether the episode has ended. The reason for recording the current entity is that we need to get all exit paths of the current entity for further fusion \ref{EQUATION-PA-MLP-1}.

\section{Experiments and Results}
\label{sec:experiments}
\subsection{Datasets}
OpenDialKG is a publicly available parallel corpus of conversations and Knowledge Graphs consisting of 91,000 conversations, each supplemented by paths connecting Knowledge Graph entities and their relationships. The purpose of the corpus is to present the implicit reasoning processes of human dialog as explicit computer operations on the Knowledge Graph. Following previous work described in \citet{moon-etal-2019-opendialkg}, we split this dataset into a 70\% training set, a 15\% validation set, and a 15\% test set.

\subsection{Baselines}
We compared our results with these baseline models: Tri-LSTM, Seq2Seq, Seq2Path, DialKG Walker\citep{Young_Cambria_Chaturvedi_Zhou_Biswas_Huang_2018, moon-etal-2019-opendialkg}, DiffKG, AttnFlow, AttnIO\citep{jung-etal-2020-attnio} and HiTKG\citep{Ni_Pandelea_Young_Zhou_Cambria_2022}. 
HiTKG is a hierarchical transformer-based tool that uses diverse inputs to predict KG paths. Our team chose HiTKG as a strong baseline. To evaluate the performance of the state-of-the-art LLMs on the KG inference task, we designed three prompt methods: GPT4-Standard, GPT4-Normal and GPT4-OPA. The difference between GPT4-Standard, GPT4-Normal is that GPT4-Normal has more awareness of dialog context and path history, while GPT4-OPA has more awareness of 2-hop exit path subgraphs compared to the former two. See appendix \ref{GPT4-Prompts} for full prompt.

\subsection{Implement Details}
The training was conducted on A40. Informed by prior research from \citet{jung-etal-2020-attnio, Ni_Pandelea_Young_Zhou_Cambria_2022}, we pre-trained the knowledge graph using TransE \citep{fan-etal-2014-transition} based on this GitHub repository\footnote{\url{https://github.com/thunlp/OpenKE}}. The objective was to unearth and explore entity relationships, expand the knowledge graph for connection prediction, and enable diverse reward function design. To facilitate reproducibility, we adopt an open-source
LLM, i.e., LLaMA-2-7B\citep{touvron2023llama2}. To reduce GPU memory usage and increase the pace of training, all experiments - excluding LLaMA-2-7B-ARK-FP32 were carried out with BFLOAT16\citep{kalamkar2019study} half-precision format. Since all true paths in OpenDialKG are at most 2 hops, we set the maximum path length to $t=2$. We included "Equal" to ensure that the model stops automatically after the second hop. To ensure fairness, we randomly shuffled the exit paths of the knowledge graph. We set max patience to 5, meaning that training is terminated if there is no boost for 5 rewards on the validation set. Further information on the hyperparameters is available in the Appendix \ref{Tab08}.

\subsection{Evaluation Metrics}
In line with the baselines, we utilize recall@k as the evaluation metric for both path-level (path@k) and target entity-level (target@k) correctness.

\subsection{Comparative Experiments}
\begin{table*}
    \centering
    \resizebox{\linewidth}{!}{
    \begin{tabular}{ccccccccccc}
    \toprule
    \multirow{3}{*}{Model} & \multicolumn{5}{c}{path@k} & \multicolumn{5}{c}{target@k} \\
    \cmidrule(r){2-6}
    \cmidrule(r){7-11}
    & path@1 & path@3  & path@5   & path@10 & path@25 & target@1 & target@3  & target@5   & target@10 & target@25  \\
    \midrule
    Tri-LSTM &3.2 &14.2 &22.6 &36.3 &56.2 &- &- &- &- &- \\
    Seq2Seq &3.1 &18.3 &29.7 &44.1 &60.2 &- &- &- &- &- \\
    DialKG Walker &13.2 &26.1 &35.3 &47.9 &62.2 &- &- &- &- &- \\
    Seq2Path     &14.92 &24.95 &31.1 &38.68 &48.15 &15.65 &27.04 &33.86 &42.52 &53.28 \\
    AttnFlow      &17.37 &24.84 &30.68 &39.48 &51.4 &18.97 &36.23 &45.48 &58.84 &71.35 \\
    AttnIO      &23.72 &37.53 &43.57 &52.17 &62.86 &24.98 &43.78 &53.49 &65.48 &78.79 \\
    HiTKG     &\textbf{25.99} &\textbf{38.67} &\textbf{49.18} &\textbf{59.32} &\textbf{71.27}& 31.11 & 46.29  & 55.59  & 71.61 & 86.09 \\
    T5-DiffKG    & - & -  & -  & - & - & 26.80 & \textbf{54.33} & 61.75  & - & - \\
    GPT-4-Standard    & 0.007 & -  & -  & - & - & 14.91 & - & - & - & - \\
    GPT-4-Normal    & 0.02 & -  & -  & - & - & 13.30 & - & - & - & - \\
    GPT-4-OPA    & 0.09 & -  & -  & - & - & 12.19 & -  & -  & - & - \\
    \midrule
    LLaMA-2-7B-ARK & 16.59 & 27.17 & 34.85  & 47.88  & 62.32 & \textbf{36.39} & 53.63  &\textbf{65.68} & \textbf{80.20} & \textbf{89.68} \\
    \bottomrule
    \end{tabular}
    }
    \caption{Path-level (path@k) and target-level (target@k) performance of KG path reasoning. LLM-ARK is benchmarked against several state-of-the-art baselines models on the OpenDialKG dataset.}
\label{Tab01}
\end{table*}

For the KG reasoning task, we assessed path recall at different K values (1, 3, 5, 10, 25) and target entity recall at position K (1, 3, 5, 10, 25). As presented in Table \ref{Tab01}, the result demonstrate that our proposed model LLaMA-2-7b-ARK performs better than all benchmarked baselines in target@1, 5, 10, 25 metrics. The performance gain is signifcant, especially in recalls with taget@1, 10: there is a 5.28\% relative improvement in target@1 and 9.59\% in target@10. Unfortunately, our model's path@k evaluation matrix socores do not outperform the current state-of-the-art (SOTA) model HiTKG because we trained using only the target arrival reward function, but we are very extensible and there is potential for improvement. As described in this paper, we evaluate the performance of GPT-4 in performing dialog inference using manual prompt constructed with different environmental information. Therefore, we also report the performance of GPT-4 with different prompts on the same dataset.

At the decoding stage of AttnIO, AttnFlow and DiffKG, KG paths are predicted by scoring entity paths and relation paths respectively, and then rerank which makes it harder to achieve optimum. While a KG triple is composed of both, our model uses PA-MLP to aggregate all the exit path information to improve the perception of the agent, which is a more reasonable modeling approach.

HiTKG is state-of-the-art KG walker, which build
a multi-hierarchy attention block to aggregate the multiscale information. However, different types of input data sources are difficult to aggregate, and how well they are aggregated directly affects the performance of the model. We unify all the multi-scale input sources into the prompt, and due to the large model has a large number of instruction comprehension ability to get a rich information encoding representation.

The GPT-4-Standard and GPT-4-Normal methods are deficient in path awareness. GPT-4-OPA exhibit improved outcomes with the addition of path awareness. The generation of GPT-4 paths is entirely dependent on the background knowledge in the dataset during the training phase, and the GPT-4 generative model itself is not designed for sequential decision-making tasks, and achieving such a score has impressed us. Although we added states to GPT-4 through prompt, this is limited by the length of the prompt, which is fundamentally due to the fact that GPT-4 is inherently memoryless. Based on these factors, optimizing GPT-4 on multi-hop inference datasets of the Knowledge Graph to further improve its performance, generating human-preferred inference paths on large-scale Knowledge Graph datasets is still a challenge.

Considering LLMs as agents that explore a knowledge graph to acquire experience can benefit benefits from the positive-negative feedback optimization mechanism of the Reinforcement Learning Policy Supervisor Algorithm. This method enhances the training of our model to perform flexible reasoning on KGs in multi-step scenarios, outperforming not only GPT-4 but also smaller models. The model's superior performance corroborates the effectiveness of our approach.

\begin{table*}
    \centering
    \resizebox{\linewidth}{!}{
    \begin{tabular}{ccccccccccc}
    \toprule
    \multirow{3}{*}{Model} & \multicolumn{5}{c}{path@k} & \multicolumn{5}{c}{target@k} \\
    \cmidrule(r){2-6}
    \cmidrule(r){7-11}
    & path@1 & path@3  & path@5   & path@10 & path@25 & target@1 & target@3  & target@5   & target@10 & target@25  \\
    \midrule
    LLaMA-2-7B-ARK & 16.59 & \textbf{27.17} & \textbf{34.85}  & \textbf{47.88}  & 62.32 & \textbf{36.39} & \textbf{53.63}  &\textbf{65.68} & \textbf{80.20} & 89.68 \\
    LLaMA-2-7B-ARK-UI & \textbf{16.87} & 27.01 & 34.35  & 47.67  & \textbf{63.03} & 34.70 & 52.02 & 62.66 & 78.09 & 88.65 \\
    LLaMA-2-7B-ARK-FP32 & 14.59 & 24.64 & 32.24  & 45.80  & 61.75 & 34.35 & 
    52.57 & 62.51 & 79.18 & 88.51  \\
    LLaMA-2-7B-ARK-WT & 1.10 & 3.44 & 5.71  & 10.47  & 15.42 & 9.45 & 
    19.46 & 50.94 & 71.43 & \textbf{94.03} \\
    LLaMA-2-7B-ARK-WP & 0.98 & 2.53 & 3.28  & 4.88  & 5.52 & 18.90 & 
    40.67 & 54.76 & 77.25 & 93.08 \\
    \bottomrule
    \end{tabular}
    }
    \caption{Path-level (path@k) and target-level (target@k) performance of supervised KG path reasoning ( metric:
recall@k). LLM-ARK is benchmarked against several ablation models on the OpenDialKG dataset.}
\label{Tab02}
\end{table*}

\begin{table*}
    \centering
    \resizebox{\linewidth}{!}{
\begin{tabular}{|clll|}
\hline
\multicolumn{4}{|l|}{\begin{tabular}[c]{@{}l@{}}\#\#\# Task Background:\\ Performing 2-hop reasoning on the knowledge graph. \\ \#\#\# Instruction\\ If you don't think it's necessary to perform the second hop in reasoning, stop the reasoning with the 'Equal' relation.\\ Given the Task Background and the Environment, directly output this path in triplet format without any other content.\end{tabular}} \\ \hline
\multicolumn{1}{|c|}{\multirow{3}{*}{Success}} &
  \multicolumn{1}{l|}{FTE} &
  \multicolumn{2}{l|}{\begin{tabular}[c]{@{}l@{}}\#\#\# Environment:\\ Dialog History: {[}{]}\\ Utterance: Could you recommend popular books by Gail Carson Levine?\\ Path History: {[}{]}\\ Current Entity: Gail Carson Levine\end{tabular}} \\ \cline{2-4} 
\multicolumn{1}{|c|}{} &
  \multicolumn{1}{l|}{Ground Truth Path} &
  \multicolumn{2}{l|}{{[}"Gail Carson Levine","$\sim$written\_by","The Two Princesses of Bamarre"{]}} \\ \cline{2-4} 
\multicolumn{1}{|c|}{} &
  \multicolumn{1}{l|}{LLM-ARK Reasoning Path} &
  \multicolumn{2}{l|}{{[}{[}"Gail Carson Levine","$\sim$written\_by","The Two Princesses of Bamarre"{]}, {[}"The Two Princesses of Bamarre","Equal","The Two Princesses of Bamarre"{]}{]}} \\ \hline
\multicolumn{1}{|c|}{\multirow{3}{*}{Failed}} &
  \multicolumn{1}{l|}{FTE} &
  \multicolumn{2}{l|}{\begin{tabular}[c]{@{}l@{}}\#\#\# Environment:\\ Dialog History: {[}"user: Can you recommend a movie like the Shooter?",\\ "assistant: A movie similar to Shooter is Nothing to Lose."{]}\\ Utterance: "Ok who is in that one?"\\ Path History: {[}{[}"Shooter","has\_genre","Thriller"{]},{[}"Thriller","$\sim$has\_genre","Nothing to Lose"{]},{[}"Nothing to Lose","starred\_actors","Michael McKean"{]}{]}\\ Current Entity: "Michael McKean"\end{tabular}} \\ \cline{2-4} 
\multicolumn{1}{|c|}{} &
  \multicolumn{1}{l|}{Ground Truth Path} &
  \multicolumn{2}{l|}{{[}"Michael McKean","$\sim$starred\_actors","Nothing to Lose"{]}} \\ \cline{2-4} 
\multicolumn{1}{|c|}{} &
  \multicolumn{1}{l|}{LLM-ARK Reasoning Path} &
  \multicolumn{2}{l|}{{[}{[}"Michael McKean","$\sim$starred\_actors","Used Cars"{]},{[}"Used Cars", "Equal", "Used Cars"{]}{]}} \\ \hline
\end{tabular}}
    \caption{Successes and failures of our model when performing inference tasks on the OpenDialKG dataset.}
\label{Tab03}
\end{table*}

\subsection{Analysis Experiment}
As shown in Table \ref{Tab02}, LLM-ARK was benchmarked against multiple ablation models on the OpenDialKG dataset. \textbf{(1)} First, to evaluate the impact of instructions on model performance, we trained the LLaMA-2-7B-ARK-UI model without instruction. The results of this model are the closest to those of LLaMA-2-7B-ARK, indicating that the presence or absence of commands has an effect on the model's results, but not a serious one. \textbf{(2)} Then, we have implemented IEEE 754 floating-point format (FP32) operations for our experiments. The results show that using the BFLOAT16 tensor for training, recall@K gives better results than FP32 without changing the hyperparameters. \textbf{(3)} Next, LLaMA-2-7B-ARK-WT that was trained by randomly initializing relation and entity embedding. The decrease in performance indicates that the absence of knowledge graph representation learning has negatively impacted the training process. \textbf{(4)} We conducted the fourth ablation experiment LLaMA-2-7B-ARK-WP and found that the performance of our export environment-aware sub-module PA-MLP decreases substantially if we do not consider the exit paths. The  score for path@1 is only 0.98\%, which is 15.61\% lower than LLaMA-2-7B-ARK. These ablation experiments and results demonstrate the contribution and necessity of our subcomponents to the model.

\subsection{Case Study}

We resort to a case study, for a clear
presentation of LLM-ARK’s path reasoning process as shown in Table \ref{Tab03}.
Note that there are hundreds of neighbor nodes connected to each entity in the external KG. Intuitively, there could be diverse knowledge paths as response to the user’s question.
As the success story shows, our model makes good use of FTE information and exit path information to make decisions, rather than making decisions based on relationships alone, because Gail Carson Levine's work is not limited to The Two Princesses of Bamarre.
As shown in the error case, our model still reasons about wrong paths, partly due to the dataset itself, because OpenDialKG is an open-domain conversational knowledge graph inference dataset, and similar contexts and the same starting entities in the training set choose different Groud Truth exit paths, and so it can interfere with the training of our model. It is worth mentioning that OpenDialKG is not a unique path inference; there are many potential paths to reach the target entity. To summarize, our model would have the potential for better performance on non-open-domain conversational knowledge graph inference datasets.

\section{Conclusion}
This paper evaluates the ability of current state-of-the-art LLM-based dialog systems in handling KG conversational reasoning tasks. To enhance LLM's performance on this task, we introduce LLM-ARK, a full-text environment-aware Knowledge Graph inference agent optimized using online reinforcement learning. Empirical analysis demonstrates that our model outperforms GPT-4 and smaller models. The experiments also sheds light on the model's performance can be seriously affected by the mismatch between the LLMs and the environment information. Our method can inspire subsequent researchers to pay attention to the critical role of considering various factors during model optimization in the field of LLM-based conversational KG reasoning tasks.

\bibliographystyle{unsrtnat}

\begin{thebibliography}{31}
\expandafter\ifx\csname natexlab\endcsname\relax\def\natexlab#1{#1}\fi

\bibitem[{Carta et~al.(2023)Carta, Romac, Wolf, Lamprier, Sigaud, and
  Oudeyer}]{carta2023grounding}
Thomas Carta, Clément Romac, Thomas Wolf, Sylvain Lamprier, Olivier Sigaud,
  and Pierre-Yves Oudeyer. 2023.
\newblock \href {http://arxiv.org/abs/2302.02662} {Grounding large language
  models in interactive environments with online reinforcement learning}.

\bibitem[{Dong et~al.(2022)Dong, Wieting, and Verga}]{dong2022faithful}
Yue Dong, John Wieting, and Pat Verga. 2022.
\newblock \href {http://arxiv.org/abs/2204.13761} {Faithful to the document or
  to the world? mitigating hallucinations via entity-linked knowledge in
  abstractive summarization}.

\bibitem[{Du et~al.(2023)Du, Watkins, Wang, Colas, Darrell, Abbeel, Gupta, and
  Andreas}]{pmlr-v202-du23f}
Yuqing Du, Olivia Watkins, Zihan Wang, C\'{e}dric Colas, Trevor Darrell, Pieter
  Abbeel, Abhishek Gupta, and Jacob Andreas. 2023.
\newblock \href {https://proceedings.mlr.press/v202/du23f.html} {Guiding
  pretraining in reinforcement learning with large language models}.
\newblock In \emph{Proceedings of the 40th International Conference on Machine
  Learning}, volume 202 of \emph{Proceedings of Machine Learning Research},
  pages 8657--8677. PMLR.

\bibitem[{Fan et~al.(2014)Fan, Zhou, Chang, and
  Zheng}]{fan-etal-2014-transition}
Miao Fan, Qiang Zhou, Emily Chang, and Thomas~Fang Zheng. 2014.
\newblock \href {https://aclanthology.org/Y14-1039} {Transition-based knowledge
  graph embedding with relational mapping properties}.
\newblock In \emph{Proceedings of the 28th Pacific Asia Conference on Language,
  Information and Computing}, pages 328--337, Phuket,Thailand. Department of
  Linguistics, Chulalongkorn University.

\bibitem[{Huang et~al.(2022)Huang, Xia, Xiao, Chan, Liang, Florence, Zeng,
  Tompson, Mordatch, Chebotar, Sermanet, Brown, Jackson, Luu, Levine, Hausman,
  and Ichter}]{huang2022inner}
Wenlong Huang, Fei Xia, Ted Xiao, Harris Chan, Jacky Liang, Pete Florence, Andy
  Zeng, Jonathan Tompson, Igor Mordatch, Yevgen Chebotar, Pierre Sermanet, Noah
  Brown, Tomas Jackson, Linda Luu, Sergey Levine, Karol Hausman, and Brian
  Ichter. 2022.
\newblock \href {http://arxiv.org/abs/2207.05608} {Inner monologue: Embodied
  reasoning through planning with language models}.

\bibitem[{Jung et~al.(2020)Jung, Son, and Lyu}]{jung-etal-2020-attnio}
Jaehun Jung, Bokyung Son, and Sungwon Lyu. 2020.
\newblock \href {https://doi.org/10.18653/v1/2020.emnlp-main.280} {{A}ttn{IO}:
  {K}nowledge {G}raph {E}xploration with {I}n-and-{O}ut {A}ttention {F}low for
  {K}nowledge-{G}rounded {D}ialogue}.
\newblock In \emph{Proceedings of the 2020 Conference on Empirical Methods in
  Natural Language Processing (EMNLP)}, pages 3484--3497, Online. Association
  for Computational Linguistics.

\bibitem[{Kalamkar et~al.(2019)Kalamkar, Mudigere, Mellempudi, Das, Banerjee,
  Avancha, Vooturi, Jammalamadaka, Huang, Yuen, Yang, Park, Heinecke,
  Georganas, Srinivasan, Kundu, Smelyanskiy, Kaul, and
  Dubey}]{kalamkar2019study}
Dhiraj Kalamkar, Dheevatsa Mudigere, Naveen Mellempudi, Dipankar Das, Kunal
  Banerjee, Sasikanth Avancha, Dharma~Teja Vooturi, Nataraj Jammalamadaka,
  Jianyu Huang, Hector Yuen, Jiyan Yang, Jongsoo Park, Alexander Heinecke,
  Evangelos Georganas, Sudarshan Srinivasan, Abhisek Kundu, Misha Smelyanskiy,
  Bharat Kaul, and Pradeep Dubey. 2019.
\newblock \href {http://arxiv.org/abs/1905.12322} {A study of bfloat16 for deep
  learning training}.

\bibitem[{Kazemi et~al.(2023)Kazemi, Kim, Bhatia, Xu, and
  Ramachandran}]{kazemi2023lambada}
Mehran Kazemi, Najoung Kim, Deepti Bhatia, Xin Xu, and Deepak Ramachandran.
  2023.
\newblock \href {http://arxiv.org/abs/2212.13894} {Lambada: Backward chaining
  for automated reasoning in natural language}.

\bibitem[{Lai et~al.(2023)Lai, Ngo, Veyseh, Man, Dernoncourt, Bui, and
  Nguyen}]{lai2023chatgpt}
Viet~Dac Lai, Nghia~Trung Ngo, Amir Pouran~Ben Veyseh, Hieu Man, Franck
  Dernoncourt, Trung Bui, and Thien~Huu Nguyen. 2023.
\newblock \href {http://arxiv.org/abs/2304.05613} {Chatgpt beyond english:
  Towards a comprehensive evaluation of large language models in multilingual
  learning}.

\bibitem[{Liu et~al.(2023)Liu, Hu, Wen, and Yu}]{liu2023comprehensive}
Aiwei Liu, Xuming Hu, Lijie Wen, and Philip~S. Yu. 2023.
\newblock \href {http://arxiv.org/abs/2303.13547} {A comprehensive evaluation
  of chatgpt's zero-shot text-to-sql capability}.

\bibitem[{Luo et~al.(2023)Luo, Li, Haffari, and Pan}]{luo2023reasoning}
Linhao Luo, Yuan-Fang Li, Gholamreza Haffari, and Shirui Pan. 2023.
\newblock \href {http://arxiv.org/abs/2310.01061} {Reasoning on graphs:
  Faithful and interpretable large language model reasoning}.

\bibitem[{Moon et~al.(2019)Moon, Shah, Kumar, and
  Subba}]{moon-etal-2019-opendialkg}
Seungwhan Moon, Pararth Shah, Anuj Kumar, and Rajen Subba. 2019.
\newblock \href {https://doi.org/10.18653/v1/P19-1081} {{O}pen{D}ial{KG}:
  Explainable conversational reasoning with attention-based walks over
  knowledge graphs}.
\newblock In \emph{Proceedings of the 57th Annual Meeting of the Association
  for Computational Linguistics}, pages 845--854, Florence, Italy. Association
  for Computational Linguistics.

\bibitem[{Ni et~al.(2022)Ni, Pandelea, Young, Zhou, and
  Cambria}]{Ni_Pandelea_Young_Zhou_Cambria_2022}
Jinjie Ni, Vlad Pandelea, Tom Young, Haicang Zhou, and Erik Cambria. 2022.
\newblock \href {https://doi.org/10.1609/aaai.v36i10.21360} {Hitkg: Towards
  goal-oriented conversations via multi-hierarchy learning}.
\newblock \emph{Proceedings of the AAAI Conference on Artificial Intelligence},
  36(10):11112--11120.

\bibitem[{OpenAI(2023)}]{openai2023gpt4}
OpenAI. 2023.
\newblock \href {http://arxiv.org/abs/2303.08774} {Gpt-4 technical report}.

\bibitem[{Ouyang et~al.(2022)Ouyang, Wu, Jiang, Almeida, Wainwright, Mishkin,
  Zhang, Agarwal, Slama, Ray, Schulman, Hilton, Kelton, Miller, Simens, Askell,
  Welinder, Christiano, Leike, and Lowe}]{Ouyang2022TrainingLM}
Long Ouyang, Jeff Wu, Xu~Jiang, Diogo Almeida, Carroll~L. Wainwright, Pamela
  Mishkin, Chong Zhang, Sandhini Agarwal, Katarina Slama, Alex Ray, John
  Schulman, Jacob Hilton, Fraser Kelton, Luke~E. Miller, Maddie Simens, Amanda
  Askell, Peter Welinder, Paul~Francis Christiano, Jan Leike, and Ryan~J. Lowe.
  2022.
\newblock \href {https://api.semanticscholar.org/CorpusID:246426909} {Training
  language models to follow instructions with human feedback}.
\newblock \emph{ArXiv}, abs/2203.02155.

\bibitem[{Peng et~al.(2023)Peng, Galley, He, Cheng, Xie, Hu, Huang, Liden, Yu,
  Chen, and Gao}]{peng2023check}
Baolin Peng, Michel Galley, Pengcheng He, Hao Cheng, Yujia Xie, Yu~Hu, Qiuyuan
  Huang, Lars Liden, Zhou Yu, Weizhu Chen, and Jianfeng Gao. 2023.
\newblock \href {http://arxiv.org/abs/2302.12813} {Check your facts and try
  again: Improving large language models with external knowledge and automated
  feedback}.

\bibitem[{Rawte et~al.(2023)Rawte, Chakraborty, Pathak, Sarkar, Tonmoy, Chadha,
  Sheth, and Das}]{rawte-etal-2023-troubling}
Vipula Rawte, Swagata Chakraborty, Agnibh Pathak, Anubhav Sarkar, S.M
  Towhidul~Islam Tonmoy, Aman Chadha, Amit Sheth, and Amitava Das. 2023.
\newblock \href {https://doi.org/10.18653/v1/2023.emnlp-main.155} {The
  troubling emergence of hallucination in large language models - an extensive
  definition, quantification, and prescriptive remediations}.
\newblock In \emph{Proceedings of the 2023 Conference on Empirical Methods in
  Natural Language Processing}, pages 2541--2573, Singapore. Association for
  Computational Linguistics.

\bibitem[{Schulman et~al.(2017)Schulman, Wolski, Dhariwal, Radford, and
  Klimov}]{DBLP:journals/corr/SchulmanWDRK17}
John Schulman, Filip Wolski, Prafulla Dhariwal, Alec Radford, and Oleg Klimov.
  2017.
\newblock \href {http://arxiv.org/abs/1707.06347} {Proximal policy optimization
  algorithms}.
\newblock \emph{CoRR}, abs/1707.06347.

\bibitem[{Shakarian et~al.(2023)Shakarian, Koyyalamudi, Ngu, and
  Mareedu}]{shakarian2023independent}
Paulo Shakarian, Abhinav Koyyalamudi, Noel Ngu, and Lakshmivihari Mareedu.
  2023.
\newblock \href {http://arxiv.org/abs/2302.13814} {An independent evaluation of
  chatgpt on mathematical word problems (mwp)}.

\bibitem[{Shinn et~al.(2023)Shinn, Cassano, Berman, Gopinath, Narasimhan, and
  Yao}]{shinn2023reflexion}
Noah Shinn, Federico Cassano, Edward Berman, Ashwin Gopinath, Karthik
  Narasimhan, and Shunyu Yao. 2023.
\newblock \href {http://arxiv.org/abs/2303.11366} {Reflexion: Language agents
  with verbal reinforcement learning}.

\bibitem[{Singh et~al.(2022)Singh, Blukis, Mousavian, Goyal, Xu, Tremblay, Fox,
  Thomason, and Garg}]{singh2022progprompt}
Ishika Singh, Valts Blukis, Arsalan Mousavian, Ankit Goyal, Danfei Xu, Jonathan
  Tremblay, Dieter Fox, Jesse Thomason, and Animesh Garg. 2022.
\newblock \href {http://arxiv.org/abs/2209.11302} {Progprompt: Generating
  situated robot task plans using large language models}.

\bibitem[{Sun et~al.(2023)Sun, Xu, Tang, Wang, Lin, Gong, Ni, Shum, and
  Guo}]{sun2023thinkongraph}
Jiashuo Sun, Chengjin Xu, Lumingyuan Tang, Saizhuo Wang, Chen Lin, Yeyun Gong,
  Lionel~M. Ni, Heung-Yeung Shum, and Jian Guo. 2023.
\newblock \href {http://arxiv.org/abs/2307.07697} {Think-on-graph: Deep and
  responsible reasoning of large language model on knowledge graph}.

\bibitem[{Touvron et~al.(2023{\natexlab{a}})Touvron, Lavril, Izacard, Martinet,
  Lachaux, Lacroix, Rozière, Goyal, Hambro, Azhar, Rodriguez, Joulin, Grave,
  and Lample}]{touvron2023llama}
Hugo Touvron, Thibaut Lavril, Gautier Izacard, Xavier Martinet, Marie-Anne
  Lachaux, Timothée Lacroix, Baptiste Rozière, Naman Goyal, Eric Hambro,
  Faisal Azhar, Aurelien Rodriguez, Armand Joulin, Edouard Grave, and Guillaume
  Lample. 2023{\natexlab{a}}.
\newblock \href {http://arxiv.org/abs/2302.13971} {Llama: Open and efficient
  foundation language models}.

\bibitem[{Touvron et~al.(2023{\natexlab{b}})Touvron, Martin, Stone, Albert,
  Almahairi, Babaei, Bashlykov, Batra, Bhargava, Bhosale, Bikel, Blecher,
  Ferrer, Chen, Cucurull, Esiobu, Fernandes, Fu, Fu, Fuller, Gao, Goswami,
  Goyal, Hartshorn, Hosseini, Hou, Inan, Kardas, Kerkez, Khabsa, Kloumann,
  Korenev, Koura, Lachaux, Lavril, Lee, Liskovich, Lu, Mao, Martinet, Mihaylov,
  Mishra, Molybog, Nie, Poulton, Reizenstein, Rungta, Saladi, Schelten, Silva,
  Smith, Subramanian, Tan, Tang, Taylor, Williams, Kuan, Xu, Yan, Zarov, Zhang,
  Fan, Kambadur, Narang, Rodriguez, Stojnic, Edunov, and
  Scialom}]{touvron2023llama2}
Hugo Touvron, Louis Martin, Kevin Stone, Peter Albert, Amjad Almahairi, Yasmine
  Babaei, Nikolay Bashlykov, Soumya Batra, Prajjwal Bhargava, Shruti Bhosale,
  Dan Bikel, Lukas Blecher, Cristian~Canton Ferrer, Moya Chen, Guillem
  Cucurull, David Esiobu, Jude Fernandes, Jeremy Fu, Wenyin Fu, Brian Fuller,
  Cynthia Gao, Vedanuj Goswami, Naman Goyal, Anthony Hartshorn, Saghar
  Hosseini, Rui Hou, Hakan Inan, Marcin Kardas, Viktor Kerkez, Madian Khabsa,
  Isabel Kloumann, Artem Korenev, Punit~Singh Koura, Marie-Anne Lachaux,
  Thibaut Lavril, Jenya Lee, Diana Liskovich, Yinghai Lu, Yuning Mao, Xavier
  Martinet, Todor Mihaylov, Pushkar Mishra, Igor Molybog, Yixin Nie, Andrew
  Poulton, Jeremy Reizenstein, Rashi Rungta, Kalyan Saladi, Alan Schelten, Ruan
  Silva, Eric~Michael Smith, Ranjan Subramanian, Xiaoqing~Ellen Tan, Binh Tang,
  Ross Taylor, Adina Williams, Jian~Xiang Kuan, Puxin Xu, Zheng Yan, Iliyan
  Zarov, Yuchen Zhang, Angela Fan, Melanie Kambadur, Sharan Narang, Aurelien
  Rodriguez, Robert Stojnic, Sergey Edunov, and Thomas Scialom.
  2023{\natexlab{b}}.
\newblock \href {http://arxiv.org/abs/2307.09288} {Llama 2: Open foundation and
  fine-tuned chat models}.

\bibitem[{Tuan et~al.(2022)Tuan, Beygi, Fazel-Zarandi, Gao, Cervone, and
  Wang}]{tuan-etal-2022-towards}
Yi-Lin Tuan, Sajjad Beygi, Maryam Fazel-Zarandi, Qiaozi Gao, Alessandra
  Cervone, and William~Yang Wang. 2022.
\newblock \href {https://doi.org/10.18653/v1/2022.findings-acl.33} {Towards
  large-scale interpretable knowledge graph reasoning for dialogue systems}.
\newblock In \emph{Findings of the Association for Computational Linguistics:
  ACL 2022}, pages 383--395, Dublin, Ireland. Association for Computational
  Linguistics.

\bibitem[{Tucker et~al.(2018)Tucker, Bhupatiraju, Gu, Turner, Ghahramani, and
  Levine}]{tucker2018mirage}
George Tucker, Surya Bhupatiraju, Shixiang Gu, Richard~E. Turner, Zoubin
  Ghahramani, and Sergey Levine. 2018.
\newblock \href {http://arxiv.org/abs/1802.10031} {The mirage of
  action-dependent baselines in reinforcement learning}.

\bibitem[{Wei et~al.(2023)Wei, Wang, Schuurmans, Bosma, Ichter, Xia, Chi, Le,
  and Zhou}]{wei2023chainofthought}
Jason Wei, Xuezhi Wang, Dale Schuurmans, Maarten Bosma, Brian Ichter, Fei Xia,
  Ed~Chi, Quoc Le, and Denny Zhou. 2023.
\newblock \href {http://arxiv.org/abs/2201.11903} {Chain-of-thought prompting
  elicits reasoning in large language models}.

\bibitem[{Yao et~al.(2023)Yao, Heinecke, Niebles, Liu, Feng, Xue, Murthy, Chen,
  Zhang, Arpit, Xu, Mui, Wang, Xiong, and Savarese}]{yao2023retroformer}
Weiran Yao, Shelby Heinecke, Juan~Carlos Niebles, Zhiwei Liu, Yihao Feng,
  Le~Xue, Rithesh Murthy, Zeyuan Chen, Jianguo Zhang, Devansh Arpit, Ran Xu,
  Phil Mui, Huan Wang, Caiming Xiong, and Silvio Savarese. 2023.
\newblock \href {http://arxiv.org/abs/2308.02151} {Retroformer: Retrospective
  large language agents with policy gradient optimization}.

\bibitem[{Young et~al.(2018)Young, Cambria, Chaturvedi, Zhou, Biswas, and
  Huang}]{Young_Cambria_Chaturvedi_Zhou_Biswas_Huang_2018}
Tom Young, Erik Cambria, Iti Chaturvedi, Hao Zhou, Subham Biswas, and Minlie
  Huang. 2018.
\newblock \href {https://doi.org/10.1609/aaai.v32i1.11923} {Augmenting
  end-to-end dialogue systems with commonsense knowledge}.
\newblock \emph{Proceedings of the AAAI Conference on Artificial Intelligence},
  32(1).

\bibitem[{Yu et~al.(2022)Yu, Velu, Vinitsky, Gao, Wang, Bayen, and
  Wu}]{yu2022the}
Chao Yu, Akash Velu, Eugene Vinitsky, Jiaxuan Gao, Yu~Wang, Alexandre Bayen,
  and Yi~Wu. 2022.
\newblock The surprising effectiveness of {PPO} in cooperative multi-agent
  games.
\newblock In \emph{Thirty-sixth Conference on Neural Information Processing
  Systems Datasets and Benchmarks Track}.

\bibitem[{Zhang et~al.(2020)Zhang, Liu, Xiong, and
  Liu}]{zhang-etal-2020-grounded}
Houyu Zhang, Zhenghao Liu, Chenyan Xiong, and Zhiyuan Liu. 2020.
\newblock \href {https://doi.org/10.18653/v1/2020.acl-main.184} {Grounded
  conversation generation as guided traverses in commonsense knowledge graphs}.
\newblock In \emph{Proceedings of the 58th Annual Meeting of the Association
  for Computational Linguistics}, pages 2031--2043, Online. Association for
  Computational Linguistics.

\end{thebibliography}

\appendix
\section{Appendix}
\label{sec:appendix}
\begin{table}
\centering
\resizebox{0.9\columnwidth}{!}{
\begin{tabular}{ccclcc}
\hline
\multicolumn{4}{c}{
Knowledge Graph} & \multicolumn{2}{c}{Dataset} \\ \hline
Entity     & Relation    & \multicolumn{2}{c}{Triplets}  & Train Data    & Test Data   \\ \hline
100,927    & 1,383       & \multicolumn{2}{c}{1,189,192} & 12,345        & 2,646       \\ \hline
\end{tabular}
}
    \caption{Detailed information about the number of knowledge graph entity-relationship triples and the number of dataset segmentation samples after processing the OpenDialKG dataset.}
\label{Tab04}
\end{table}

\subsection{Data Format}
We preprocessed the OpenDialKG raw data to fit our KG inference task. There are individual errors in the raw data, and the information of the dataset after our screening is shown in Table \ref{Tab04}.

\subsection{Tricks}
In the original paper of PPO, no implementation details and techniques are mentioned other than the use of GAE to compute the dominance function. Referring to this repository\footnote{\url{https://github.com/Lizhi-sjtu/DRL-code-pytorch}}, we employ several optimization tricks. In the actual code implementation, to encourage the diversity of paths sampled by the strategy during training, we added an entropy regularization term to our loss function. We used the operation of normalization of advantage proposed in the paper \citep{tucker2018mirage}. Learning rate decay can enhance the smoothness in the late training stage to some extent and improve the training effect. Here we use the linear decay of learning rate, with the number of training steps learning rate from the initial value of a linear decline to 0. Gradient clipping is a trick introduced to prevent the gradient from exploding during the training process, which also serves to stabilize the training process. Orthogonal Initialization is a neural network initialization method proposed to prevent problems such as gradient vanishing and gradient explosion at the beginning of training. Referring to the MAPPO \citep{yu2022the}, the Adam optimizer individually sets eps=1e-5, and this particular setting can improve the training performance of the algorithm to some extent.

\subsection{Limitation}
\subsubsection{Limitations of Inference Efficiency}
Efficiency is always a significant issue when building deep learning models based on LLMs. Although our research freezes the parameters of the LLM in the back-propagation stage and uniformly uses the bfloat16 computational type, the huge number of parameters of the model leads to inefficient forward propagation and large GPU memory usage when collecting experience and inference. As stated in the LLaMA paper\citep{touvron2023llama}, the efficiency of the model's inference is more crucial than its training efficiency. It is acceptable for the training process to be slower, but the inference must be faster. Improving the inference speed of the model while ensuring its effectiveness is a challenge. In addition to constructing the research model, online applications based on LLMs must also address the efficiency issue. Conversational reasoning models based on LLMs must be efficient for real-time applications. The inference efficiency is crucial for building online applications based on LLMs.

\subsubsection{Limitations of Entity Embedding}
Our research work has identified limitations in the semantic representation of knowledge graph entities. The attributes of knowledge graph entities should be considered during the reasoning process. However, these attributes may be lengthy descriptions that are not easily processed by our TransE knowledge graph semantic embedding model. Furthermore, while most knowledge graphs are currently represented in text form, it is equally important to consider multimodal knowledge graph reasoning in research. By constructing a reasoning model based on multimodal inputs, machines can better describe and understand the real world. For example, the soon-to-be two-dimensionalized Law in Three Body Death Forever says "Oh, it's time to go into the picture, kids, go ahead," and the user asks a question about this scenario, "can you help me find some pictures related to this galaxy?". The model may need to deduce that the two-dimensional representation portrayed in this book is the Milky Way galaxy, and then locate relevant images of the galaxy. Our model is currently unable to incorporate the combination of multi-modal, multi-attribute entities, which is a limitation of our work in this endeavor, as well as an area for future research efforts.

\subsection{GPT4 Prompts}
\label{GPT4-Prompts}
For GPT4-OPA prompt, since max length is 2, we need to recursively get all exit paths at the next level of all exit paths of the current entity, most of which are omitted due to the large number of KG subgraph triples of exit paths.

\begin{table}
\centering
\resizebox{0.9\columnwidth}{!}{
\begin{tabular}{|l|}
\hline
\multicolumn{1}{|c|}{Standard Prompt} \\ \hline
\begin{tabular}[c]{@{}l@{}}\#\#\# Task Background \\ Performing 2-hop reasoning on the knowledge graph. \\ \\ \#\#\# Instruction \\ If you don’t think it’s necessary to perform the second hop in reasoning, stop the reasoning with the ’Equal’ relation. \\ Given the Task Background and the Environment, directly output this path in triplet format without any other content. \\ \\ \#\#\# Environment \\ Utterance: What do you think about the Washinton Redskins? Are you a fan? 
\\ Current Entity: Washington Redskins \\ \\ \#\#\# Examples \\ · \\ · \\ · \\ \\ \#\#\# Response\end{tabular} \\ \hline
\end{tabular}
}
\caption{GPT4-Standard prompt only perceived user's query and Current Entity.}
\label{Tab05}
\end{table}

\begin{table}[!ht]
\centering
\resizebox{0.9\columnwidth}{!}{
\begin{tabular}{|l|}
\hline
\multicolumn{1}{|c|}{Normal Prompt} \\ \hline
\begin{tabular}[c]{@{}l@{}}\#\#\# Task Background \\ Performing 2-hop reasoning on the knowledge graph. \\ \\ \#\#\# Instruction \\ If you don’t think it’s necessary to perform the second hop in reasoning, stop the reasoning with the ’Equal’ relation. \\ Given the Task Background and the Environment, directly output this path in triplet format without any other content. \\ \\ \#\#\# Environment \\ Dialog History: {[}{]} \\ Utterance: What do you think about the Washinton Redskins? Are you a fan? \\ Path History: {[}{]} \\ Current Entity: Washington Redskins \\ \\ \#\#\# Examples \\ · \\ · \\ · \\ \\ \#\#\# Response\end{tabular} \\ \hline
\end{tabular}
}
\caption{GPT4-Normal prompt has more awareness of dialog context and path history.}
\label{Tab06}
\end{table}

\begin{table}[!ht]
\centering
\resizebox{0.9\columnwidth}{!}{
\begin{tabular}{|l|}
\hline
\multicolumn{1}{|c|}{OPA(Out Paths Aware) Prompt} \\ \hline
\begin{tabular}[c]{@{}l@{}}\#\#\# Task Background \\ Performing 2-hop reasoning on the knowledge graph.\\ \\ \#\#\# Instruction \\ Given the Task Background and the Environment, please choose select two consecutive paths KG path from a set of Out Paths. \\ If you don’t think it’s necessary to perform the second hop in reasoning, just select the ’Equal’ relation at the second hop. \\ Directly output these path in triplet format without any other content. \\ \\ \#\#\# Environment \\ Dialog History: {[}{]} \\ Utterance: What do you think about the Washinton Redskins? Are you a fan? \\ Path History: {[}{]} \\ Current Entity: Washington Redskins \\ Out Path: {[}'Washington Redskins,Equal,\\ Washington Redskins', \\ 'Washington Redskins,$\sim$Game,Mike Sellers', \\ 'Washington Redskins,$\sim$Runner-up,Super Bowl VII', \\ · \\ · \\ · \\ 'Ladell Betts,Ethnicity,African American'{]} \\ \\ \#\#\# Examples\\ · \\ · \\ · \\ \\ \#\#\# Response\end{tabular} \\ \hline
\end{tabular}
}
\caption{GPT4-OPA prompt has more awareness of 2-hop exit KG path subgraphs.}
\label{Tab07}
\end{table}

\subsection{HyperParameters}

\begin{table}
\centering
 \resizebox{0.9\columnwidth}{!}{
	\begin{tabular}{ll}
		\hline
		Computing Infrastructure                     & Tesla A40 GPU      \\
		Search Strategy                              & Beam Search        \\
		Training Efficiency                          & 6 seconds per step \\ \hline
		                                             &                    \\ \hline
		Hyperparameter                               & Best Setting       \\ \hline
		use transe                                   & True               \\
		out path aware                               & True               \\
		bf16                                         & True               \\
		relation embedding size                      & 200                \\
		entity embedding size                        & 200                \\
		max out                                      & 50                 \\
		number of explorations                        &  8                  \\
		replay buffer size                              & 4096               \\
		mini batch size                              & 1024              \\
		positive reward                              &  1                  \\
		negative reward                              & -1                 \\
		actor learning rate                          & 5e-5               \\
		critic learning rate                         & 5e-5               \\
		gamma                                        & 0.95               \\
		lamda                                        & 0.95               \\
		epsilon                                      & 0.2                \\
		K epochs                                     & 10                 \\
		use advantage normalization                  & True               \\
		use entropy coef                             & 0.01               \\
		use learning rate decay                      & True               \\
		use gradient clip                            & True               \\
		use orthogonal init                          & True               \\
		set adam eps 1e-5                            & True               \\
		use tanh                                     & True              
	\end{tabular}
 }
	\caption{Additional implementation detail of LLM-ARK.}
\label{Tab08}
\end{table}

\end{document}